\pdfoutput=1

\documentclass[11pt]{article}

\usepackage[preprint]{acl}

\usepackage{times}
\usepackage{latexsym}

\usepackage[T1]{fontenc}

\usepackage[utf8]{inputenc}

\usepackage{microtype}

\usepackage{inconsolata}

\usepackage{graphicx}

\usepackage{multirow}

\title{Token and Span Classification for Entity Recognition\\in French Historical Encyclopedias}

\author{
   \textbf{Ludovic Moncla} \and \textbf{Hédi Zeghidi} 
\\
\\
  INSA Lyon, CNRS, Universite Claude Bernard Lyon 1,\\LIRIS, UMR5205, 69621 Villeurbanne, France 
\\
  \small{
    \textbf{Correspondence:} \href{mailto:ludovic.moncla@insa-lyon.fr}{ludovic.moncla@insa-lyon.fr}
  }
}

\begin{document}

\maketitle

\begin{abstract}
Named Entity Recognition (NER) in historical texts presents unique challenges due to non-standardized language, archaic orthography, and nested or overlapping entities. This study benchmarks a diverse set of NER approaches, ranging from classical Conditional Random Fields (CRFs) and spaCy-based models to transformer-based architectures such as CamemBERT and sequence-labeling models like Flair. Experiments are conducted on the GeoEDdA dataset, a richly annotated corpus derived from 18th-century French encyclopedias. We propose framing NER as both token-level and span-level classification to accommodate complex nested entity structures typical of historical documents. Additionally, we evaluate the emerging potential of few-shot prompting with generative language models for low-resource scenarios. Our results demonstrate that while transformer-based models achieve state-of-the-art performance, especially on nested entities, generative models offer promising alternatives when labeled data are scarce. The study highlights ongoing challenges in historical NER and suggests avenues for hybrid approaches combining symbolic and neural methods to better capture the intricacies of early modern French text.
\end{abstract}

\section{Introduction}

The field of digital humanities increasingly depends on computational tools to extract, analyze, and structure information from vast historical corpora. Among the most significant artifacts of Enlightenment knowledge dissemination is the Encyclopédie, edited by Diderot, d’Alembert, and their collaborators. Published in the mid-18th century, this extensive compendium reflects the intellectual, geographical, and scientific worldview of the era. Extracting semantic data from such texts opens up novel opportunities for historical analysis and spatial humanities.

However, processing French historical texts poses unique challenges. Unlike modern corpora, these documents feature non-standardized language, historical orthography, and complex syntactic patterns, which hinder the straightforward application of standard Natural Language Processing (NLP) techniques \cite{mcdonough2019named}. Furthermore, geographical and named entity mentions are often embedded within hierarchical or nested contexts \cite{gaio2017extended}. These difficulties necessitate flexible and robust Named Entity Recognition (NER) methodologies tailored for historical data.

In this paper, we address these challenges by framing NER as both token-level and span-level classification tasks, capable of recognizing nested and overlapping entities, including proper nouns, common nouns, and spatial relations. 
We conduct a comprehensive benchmark of six distinct NER architectures, from classical CRF-based models and spaCy’s CNN and span-based systems to cutting-edge transformer models like CamemBERT and generative large language models (LLMs) such as GPT variants, fine-tuned or prompted on a carefully annotated subset of the Encyclopédie. Our contributions include quantitative analyses of performance on flat and nested entities, a critical comparison of few-shot learning versus fully supervised methods, and a discussion of the implications for digital humanities applications.

\section{Related works}
\label{related_works}
Named Entity Recognition (NER) has been a longstanding task in Natural Language Processing (NLP), with a rich literature base spanning over two decades. Classical NER focused on identifying entities such as persons, organizations, and locations in newswire texts, leading to the development of early rule-based systems and statistical models. Over time, the field evolved with advances in machine learning and deep learning architectures.

Rule-based NER systems were among the earliest approaches, relying on handcrafted linguistic rules and gazetteers. Although limited in adaptability, these systems offer interpretability and can be fine-tuned with domain expertise. Perdido \cite{moncla2023perdido} exemplifies a domain-specific rule-based system for French geoparsing. It performs two major tasks—geotagging (identifying spatial entities) and geocoding (resolving coordinates)—and has proven effective in French historical corpora.

Machine learning methods, particularly Conditional Random Fields (CRFs), brought probabilistic modeling to NER. CRFs consider both word-level features and contextual dependencies, enabling globally coherent label sequences. Unlike token-level classifiers that make independent predictions, CRFs model the conditional probability of an entire label sequence given the input, which is crucial for ensuring valid tag transitions. When combined with contextualized representations from models like BiLSTMs or transformers, CRFs provide a mechanism for enforcing global consistency in label assignments, leading to improved performance on benchmark datasets \cite{huang2015bidirectional}.

Building on the success of CRFs and neural architectures, Flair \cite{akbik2019flair} integrates character-level language model embeddings with a BiLSTM-CRF sequence labeling architecture. Flair embeddings capture subword information and contextual dependencies more effectively than traditional word embeddings. The framework supports various embeddings—including contextual string embeddings, BERT, and ELMo—which can be stacked to enhance model robustness. The underlying BiLSTM captures contextual information bidirectionally, while the CRF layer models dependencies between output labels to ensure globally optimal tag sequences.

Transformer-based models, particularly BERT (Bidirectional Encoder Representations from Transformers) \cite{devlin2018bert}, and its French variant CamemBERT \cite{martin2020camembert}, have significantly advanced the state of the art in NER. BERT leverages self-attention mechanisms \cite{vaswani2017attention} to capture deep bidirectional context across entire input sequences. This enables a more nuanced understanding of word meaning based on context, which is especially beneficial for disambiguating entity types. In NER tasks, BERT is typically fine-tuned by adding a token-level classification head. This paradigm has consistently outperformed traditional architectures, including BiLSTM-CRF models, on datasets such as CoNLL-2003 and OntoNotes 5.0.

Building upon BERT’s architecture, SpanBERT \cite{joshi2020spanbert} improves pretraining by focusing on span-level representations. It replaces BERT’s token-level masking with span-level masking and introduces a span-boundary objective that trains the model to predict masked spans using only the boundary token representations. This approach enables SpanBERT to learn richer contextual representations of text spans. While it has shown notable improvements on tasks like question answering and coreference resolution, its span-centric modeling is also well suited to NER, particularly for capturing multi-token entities and complex contextual patterns.

Recently, \cite{tafer2025extracting} proposed a comparative study of three architectures (including Bi-LSTM+CRF, BERT, and GliNER \cite{zaratiana2024gliner}) for extracting spatial information from itinerary descriptions in French. The annotation schema also includes nominal mentions, named entities and nested entities but applied to contemporary texts focused on displacement descriptions.

Recent studies explore prompt engineering and context scaling strategies in LLMs for NER, revealing trade-offs in performance depending on model size, example formatting, and input length \cite{xie2023empirical, li2023far, agarwal2024many, gonzalez2023yes, zeghidi2024evaluating}.

Standard NER systems typically assume flat, non-overlapping entity spans. However, texts, particularly in digital humanities, often contain overlapping and nested entities that violate this assumption. Some approaches, such as joint-labeling strategies, attempt to handle nesting by converting nested entities into combinatorial labels, reducing the task to token classification. In contrast, models like spaCy’s spancat adopt a span-based formulation, using span-entity compatibility scoring to identify and classify candidate spans. These innovations are particularly important for entity recognition in rich historical corpora, where complex entity structures are common. In this work, we build on these insights by providing an empirical comparison tailored to the domain-specific task of historical entity recognition, leveraging diverse modeling strategies and detailed evaluation protocols.

\begin{figure*}
    \centering
    \includegraphics[width=.95\linewidth]{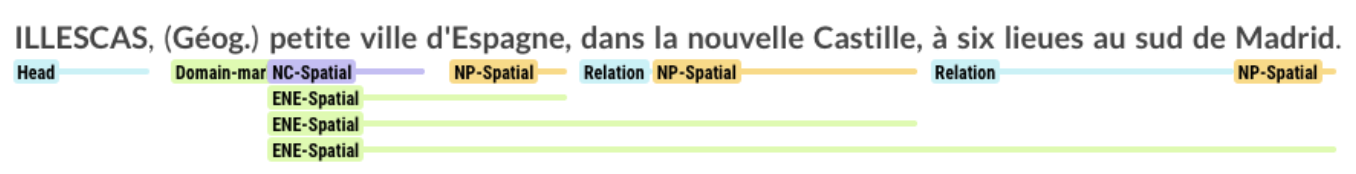}
    \caption{Excerpt of annotations from the ILLESCAS entry.}
    \label{fig:illescas}
\end{figure*}

\section{Methodology}
\label{methodology}

\subsection{Dataset Description}

The foundation of our experiments is the GeoEDdA dataset\footnote{https://huggingface.co/datasets/GEODE/GeoEDdA} \cite{moncla2024geoedda}, developed as part of the GEODE project\footnote{https://geode-project.github.io}. GeoEDdA consists of 2,200 paragraphs sampled from more than 2,000 entries in the 18th-century Encyclopédie. Each paragraph is annotated with spans corresponding to twelve different entity types, including common and proper nouns for spatial, person, and miscellaneous entities. The annotations were manually refined using active learning, leveraging the Prodigy annotation tool integrated with spaCy models \cite{moncla2024geoedda}.

Each annotation includes not only the start and end character offsets and the corresponding entity label but also contextual metadata such as paragraph source, volume number, and thematic classification (e.g., Geography, History, Medicine) \cite{brenon2022classifying}. The dataset is partitioned into three splits: 1,800 paragraphs for training, 200 for validation, and 200 for testing.
It supports nested entities by assigning overlapping spans, making it suitable for both flat and hierarchical entity recognition tasks (see Figure~\ref{fig:illescas}).

\begin{table}[ht]
\centering
 \small
\begin{tabular}{lccc}\hline
            & \textbf{Train}     & \textbf{Validation} & \textbf{Test}     \\ \hline
NC-Spatial  & 3,252     & 358        & 355      \\ 
NP-Spatial  & 4,707     & 464        & 519      \\ 
ENE-Spatial & 3,033     & 326        & 334      \\ 
Relation    & 2,093     & 219        & 226      \\ 
Latlong     & 553       & 66         & 72       \\ 
NC-Person   & 1,378     & 132        & 133      \\ 
NP-Person   & 1,599     & 170        & 150      \\ 
ENE-Person  & 492       & 49         & 57       \\ 
NP-Misc     & 948       & 108        & 96       \\ 
ENE-Misc    & 255       & 31         & 22       \\ 
Head        & 1,261     & 142        & 153      \\ 
Domain-Mark & 1,069     & 122        & 133      \\ \hline
Paragraphs  & 1,800     & 200        & 200      \\ 
Tokens      & 132,398   & 14,959     & 13,881   \\ \hline
\end{tabular}
\caption{Distribution of entity across the different sets}
\label{table:distribution_entity_sets}
\end{table}

Table~\ref{table:distribution_entity_sets} shows the distribution of entity accross the datasets. `NP` prefix is used for named entities (i.e., proper nouns), `NC` prefix for nominal mentions (i.e., common nouns) and `ENE` prefix for nested entities.
The class distribution is highly imbalanced, with spatial entities being more frequent than person or miscellaneous entities. For example, `NP-Spatial` occurs more than 4,700 times, while miscellaneous nested entities (i.e., ENE-Misc) have less than 300 occurrences.

\begin{table*}[h!]
\resizebox{\textwidth}{!}{%
\begin{tabular}{lllllllll}
\hline
    & \textbf{Domain mark}& \textbf{Head}   & \textbf{Relation}   & \textbf{NC-Person} & \textbf{NC-Spatial} & \textbf{NP-Misc}  & \textbf{NP-Person}   & \textbf{NP-Spatial} \\ \hline
Best Features  & \begin{tabular}[c]{@{}l@{}}'Token:.'\\  'lower:.'\\ 'Token:terme'\\  'lower:terme'\\ 'Token:Géograph'\end{tabular} & \begin{tabular}[c]{@{}l@{}}'prev\_Token:*'\\ 'prev\_lower:*'\\ 'Token:)'\\ 'lower:)'\\ 'isupper'\end{tabular}                       & \begin{tabular}[c]{@{}l@{}}'lower:midi'\\ 'Token:dessous'\\ 'lower:dessous'\\ 'Token:Midi'\\ 'next\_Token:petits'\end{tabular} & \begin{tabular}[c]{@{}l@{}}'lower:pape'\\ 'lower:roi'\\ 'Token:Mans'\\ 'lower:mans'\\ 'lower:président'\end{tabular}  & \begin{tabular}[c]{@{}l@{}}'lower:royaume'\\ 'lower:fleuve'\\  'lower:comté'\\  'Token:île'\\  'lower:île'\end{tabular} & \begin{tabular}[c]{@{}l@{}}'shape:dddd'\\ 'Token:persan'\\ 'lower:persan'\\ 'prev\_Token:xiij'\\ 'prev\_lower:xiij'\end{tabular} & \begin{tabular}[c]{@{}l@{}}'prev\_Token:Hunauld'\\ 'prev\_lower:hunauld'\\ 'lower:juifs'\\ 'Token:bazanés'\\ 'lower:bazanés'\end{tabular}  & \begin{tabular}[c]{@{}l@{}}'Token:ltalie'\\  'lower:ltalie'\\  'prev\_lower:palus'\\  'prev\_Token:palus'\\  'lower:indes'\end{tabular}   \\ \hline
Worst Features & \begin{tabular}[c]{@{}l@{}}'shape:Xxxxxxxxxxx'\\  'dep:appos'\\  'pos:DET'\\  'pos:PUNCT'\\ 'isupper'\end{tabular} & \begin{tabular}[c]{@{}l@{}}'prev\_lower:)'\\ 'prev\_shape:X.'\\ 'prev\_pos:NOUN'\\ 'prev\_pos:PUNCT'\\ 'prev\_shape:X'\end{tabular} & \begin{tabular}[c]{@{}l@{}}'next\_lower:se'\\ 'next\_Token:sur'\\ 'next\_lower:sur'\\ 'shape:x.'\\ 'pos:PROPN'\end{tabular}    & \begin{tabular}[c]{@{}l@{}}'shape:Xxxxxxxxxx'\\ 'isstop'\\ 'pos:ADV'\\ 'shape:Xxxxx'\\ 'next\_dep:xcomp'\end{tabular} & \begin{tabular}[c]{@{}l@{}}'next\_dep:obl:agent'\\ 'Token:du'\\ 'lower:du'\\ 'next\_dep:det'\\ 'isstop'\end{tabular}    & \begin{tabular}[c]{@{}l@{}}'prev\_Token:v.'\\ 'next\_shape:dddd'\\ 'pos:ADP'\\ 'shape:dd'\\ 'isstop'\end{tabular}                & \begin{tabular}[c]{@{}l@{}}'prev\_shape:Xxxxxxxxx'\\ 'shape:xxxxxx'\\ 'shape:xxxx'\\ 'shape:xxxxxxxxxxx'\\ 'shape:xxxxxxxxxx'\end{tabular} & \begin{tabular}[c]{@{}l@{}}'shape:xxx'\\  'next\_dep:flat:name'\\ 'shape:xxxxxxxx'\\  'shape:xxxxxxx'\\  'shape:xxxxxxxxxxx'\end{tabular} \\ \hline
\end{tabular}}
\caption{Top 5 best and worst features of CRF Model (configuration 3)}
\label{table:features_pos_dep_CRF_model}
\end{table*}

\subsection{Model Architectures and Task Formulation}

In this study, we evaluate the performance of a diverse set of NER architectures, encompassing both traditional machine learning and modern deep learning approaches. We begin with CRFs, a classical probabilistic sequence labeling method well-suited for NER. We then explore Convolutional Neural Networks (CNNs), implemented via the spaCy library, which capture local context through sliding window filters. Bidirectional Long Short-Term Memory networks (Bi-LSTMs), trained using the Flair framework \cite{akbik2019flair}, enable the modeling of sequential dependencies in both forward and backward directions, offering richer contextual representations.
To leverage recent advances in language modeling, we also fine-tune CamemBERT \cite{martin2020camembert}, a French-adapted transformer architecture based on RoBERTa, which encodes deep bidirectional context through self-attention mechanisms. Finally, we assess the potential of generative LLMs, particularly OpenAI GPTs, for structured NER tasks in a few-shot learning setup.

The GeoEDdA annotation schema supports three levels of entity nesting: flat (level 0), shallow nested (level 1), and deep nested (level 2 and beyond). In this study, we limit our scope to the flat and shallow nested levels, which already present considerable challenges due to token overlap and label ambiguity.

An illustrative example is the phrase “ville d'Espagne,” where the tokens “ville” and “Espagne” are annotated respectively as `NC-Spatial` and `NP-Spatial`, while the full phrase is labeled as `ENE-Spatial` (see Figure~\ref{fig:illescas}. This results in overlapping annotations for the same span, requiring a multi-label classification strategy.

To address this, we simplify the task by employing a joint-label encoding strategy. Overlapping entities are merged into composite labels (e.g., “NP-Spatial+ENE-Spatial”), allowing the problem to be recast as a flat classification task. These joint labels are mapped to unique class identifiers, enabling compatibility with models that do not natively support nested span prediction.

\section{Experiments and Results}
\label{experiments}

\subsection{CRF Model}

The CRF model served as a baseline for traditional sequence labeling. We explored three feature configurations: (1) basic lexical features (token, lowercase, punctuation), (2) additional syntactic features (POS tags), and (3) full features including dependency parses. Grid search over regularization parameters (C1 and C2), learning rate and dropout was conducted to mitigate overfitting. The model performed well on frequently occurring entity types like `NP-Spatial` and `Relation` but struggled on `NC-Person` and `NP-Misc` due to feature sparsity and lack of contextual depth. Results also shows that adding additional features such as part-of-speech in the feature configurations have only a very limited impact on the performances (see Table~\ref{table:crf_scores}).

\begin{table}[ht]

    \centering
     \small
    \begin{tabular}{lcccc}
        \hline
            & (1)   & (2)       & (3)       \\ \hline

NC-Spatial  & 90.0  & 89.8      & 89.6      \\ 
NP-Spatial  & 90.7  & 90.9      & 91.1      \\ 
ENE-Spatial & 89.6  & 89.9      & 89.0      \\
Relation    & 93.1  & 92.7      & 91.3      \\ 
Latlong     & 97.0  & 96.9      & 91.0      \\

NC-Person   & 58.7  & 64.0      & 62.6      \\ 
NP-Person   & 77.7  & 79.0      & 76.4      \\
ENE-Person  & 70.3  & 73.3      & 72.6      \\

NP-Misc     & 62.0  & 67.1      & 65.2      \\ 
ENE-Misc    & 43.0  & 43.9      & 40.0      \\

Head        & 86.8  & 88.1      & 85.3      \\ 
Domain-mark & 99.0  & 99.0      & 99.0      \\ 
\hline
Micro Avg   & 88.1  & 88.6      & 87.7      \\ 
Macro Avg   & 79.8  & 81.2      & 79.9      \\\hline
    \end{tabular}
    \caption{CRF F1-scores on test set with (1) basic lexical features, (2) additional syntactic features, and (3) full features.}
    \label{table:crf_scores}
\end{table}

Moreover, adding dependency parsing as a feature has a negative impact. 
Table~\ref{table:features_pos_dep_CRF_model} shows the best and worst features for configuration 3 (i.e., including part-of-speech and dependency parsing). Dependency features such as dep:appos and pos:PUNCT were among the least informative, suggesting that shallow syntactic features may introduce noise rather than signal in historical text NER.

\subsection{spaCy NER and Spancat fine-tuning}

We compare two different approaches from the spaCy platform\footnote{\url{https://spacy.io/}}: spaCy NER and spaCy Spancat\footnote{\url{https://spacy.io/api/spancategorizer}}. spaCy NER is a transition-based parser with a CNN-based model (token-based classification) while spaCy Spancat is a feed-forward neural network over span representations (span-based classification).
Table~\ref{table:spacy} shows the precision and recall scores of the two fine-tuned spaCy models for each class.

Despite rapid convergence during training, the CNN suffered from underperformance (below 60\%) on rare and nested classes (see Table~\ref{table:spacy}).
The spaCy NER architecture was more robust in recognizing contiguous (at least for precision), flat spans but lacked mechanisms for long-range dependencies and hierarchy. The overall scores also show that spaCy NER is above the CRF scores for every label (see Table~\ref{table:summary_results}).
In average Spancat has a higher recall than spaCy NER, but spacy NER has a higher precision. Except for some classes such as nested entities (`ENE-Spatial` and `ENE-Person`) and also under-represented classes such as `NP-Misc` (support from the test set is listed in Table~\ref{table:distribution_entity_sets}).

Our fine-tuned spaCy spancat model \cite{moncla2024geoedda} is available on the HuggingFace hub\footnote{\url{https://huggingface.co/GEODE/fr_spacy_custom_spancat_edda}}.

\begin{table}[t]
    \centering
    \small
    \begin{tabular}{lcccc}
        \hline
            & \multicolumn{2}{c}{\textbf{spaCy NER}} & \multicolumn{2}{c}{\textbf{Spancat}}  \\ \hline
            & \textbf{P} & \textbf{R} & \textbf{P} & \textbf{R}  \\ \hline
NC-Spatial  & \textbf{79.7}  & 34.5      & 40.9  & \textbf{59.8}   \\ 
NP-Spatial  & \textbf{93.3}  & 50.3      & 92.0   & \textbf{69.1}  \\ 
ENE-Spatial & 59.1  & \textbf{36.1}      & \textbf{86.7}  &  16.2   \\
Relation    & \textbf{62.5}  & \textbf{92.7}      & 26.9  &  69.7  \\ 
Latlong     & \textbf{92.5}  & 40.8      & 92.3  &  \textbf{86.6}   \\

NC-Person   & \textbf{82.1}  & 14.2      & 36.4  & \textbf{52.0}    \\ 
NP-Person   & 74.8  & 40.9      & \textbf{78.3}  & \textbf{69.5}    \\
ENE-Person  & 45.5  & \textbf{2.5}      & \textbf{66.7}   & 1.0   \\

NP-Misc     & 18.8  & 10.9      & \textbf{24.6}  & \textbf{34.9}    \\ 
ENE-Misc    & \textbf{2.9}  & \textbf{1.2}      & 0.00  & 0.00     \\

Head        & \textbf{92.3}  & 52.0      & 82.8  & \textbf{60.6}     \\ 
Domain-mark & \textbf{100}  & 41.1      & 92.6   & \textbf{88.8}  \\ 
\hline
Micro Avg   & \textbf{77.5}  & 33.4      & 58.5  & \textbf{57.4}    \\ 
Macro Avg   & \textbf{66.9}  & 28.6      & 60.0  & \textbf{50.7}    \\\hline
    \end{tabular}
    \caption{Precision and recall scores evaluation for our fine-tuned spaCy models.}
    \label{table:spacy}
\end{table}

\subsection{CamemBERT fine-tuning}

To evaluate the capabilities of transformer-based models on historical NER, we fine-tuned CamemBERT, a pretrained RoBERTa-like model for French \cite{martin2020camembert}, using the GeoEDdA corpus. We experimented with two output strategies for handling the overlapping and nested nature of annotations: joint-label classification and multi-label classification (span-based annotations).

As described in Section~\ref{methodology}, in the joint-label classification setup, each token is assigned a single composite label representing the combination of all entity types to which it belongs. This transforms the multi-label problem into a single-label task with an expanded label set. The multi-label classification strategy, on the other hand, models each entity type independently using a sigmoid activation for each class, allowing tokens to be tagged with multiple labels simultaneously.

As shown in Table~\ref{table:bert_scores}, both strategies achieved closely matched results, with complementary strengths. The joint-label model showed slightly better precision, producing more accurate predictions with fewer false positives. This suggests that the joint-labeling approach may help CamemBERT better model interdependencies between overlapping entity types and avoid overprediction. Conversely, the multi-label model yielded slightly higher recall, capturing a broader range of correct entities, including some low-frequency or ambiguous cases that the joint-label model missed.

These tendencies were particularly evident in complex constructions involving nested or overlapping entities, such as \textit{ville d'Espagne}, where the multi-label approach sometimes detected more entity spans but occasionally introduced inconsistencies between related labels.

Despite their differences, both strategies demonstrated strong overall performance.
Our fine-tuned CamemBERT models are available on the HuggingFace Hub\footnote{\url{https://huggingface.co/GEODE/}}.

\begin{table}[t]
    \centering
     \small
    \begin{tabular}{lcccc}
        \hline
            & \multicolumn{2}{c}{\textbf{Joint-label}} & \multicolumn{2}{c}{\textbf{Multi-label}}  \\ \hline
            & \textbf{P}       & \textbf{R}       & \textbf{P}    & \textbf{R}      \\ \hline
NC-Spatial  &  90.5            & 95.1             & \textbf{91.6} & \textbf{95.3}   \\ 
NP-Spatial  &  95.2            & \textbf{95.7}    & \textbf{95.6} &  93.7           \\ 
ENE-Spatial &  94.0            & 89.7             & \textbf{94.4} & \textbf{90.9}   \\
Relation    &  87.5            & 93.1             & \textbf{90.8} & \textbf{93.6}   \\ 
Latlong     &  \textbf{98.1}   & 96.8             & 96.0          & \textbf{98.0}    \\

NC-Person   & \textbf{67.2}    & 80.9             & 63.0          & \textbf{83.1}   \\ 
NP-Person   & \textbf{87.9}    & 86.2             & 86.1          & \textbf{88.7}   \\
ENE-Person  & \textbf{89.5}    & 77.4             & 88.7          & \textbf{86.4}   \\

NP-Misc     & 68.7             & \textbf{76.6}    & \textbf{69.6} & 76.0            \\ 
ENE-Misc    & 40.0             & 66.7             & \textbf{40.7} & \textbf{72.8}   \\

Head        & \textbf{97.3}    & \textbf{98.0}    & 97.2          & 96.5            \\ 
Domain-mark & \textbf{99.7}    & \textbf{99.0}    & 99.2          & \textbf{99.0}   \\ 
\hline
Micro Avg   &  \textbf{90.1}   & 92.0             & 89.8          & \textbf{92.7}    \\ 
Macro Avg   &  \textbf{84.6}   & 87.9             & 84.4          & \textbf{89.5}    \\\hline
    \end{tabular}
    \caption{Precision and recall scores evaluation for our fine-tuned CamemBERT models.}
    \label{table:bert_scores}
\end{table}

\subsection{Flair fine-tuning}

In addition to transformer-based approaches, we fine-tuned the Flair sequence labeling model \cite{akbik2019flair}, which is based on a BiLSTM-CRF architecture with contextual string embeddings. Flair has demonstrated strong performance on standard NER benchmarks and offers a lightweight alternative to transformer-based models, making it a relevant baseline for resource-constrained settings or historical text processing.

For our experiments, we used pretrained French Flair embeddings trained on OSCAR and Common Crawl corpora. 
Overall, Flair achieved competitive performance, especially considering its smaller size and lighter computational requirements compared to transformer models. However, its results remained slightly behind those of CamemBERT in terms of F1 score, for complex or overlapping entities (e.g., `ENE-Spatial`, `ENE-Person`, `ENE-Misc`). This is likely due to Flair's limited capacity to model long-range dependencies and inter-label interactions compared to transformer-based architectures.

\begin{table}[t]
    \centering
     \small
    \begin{tabular}{lccc}
        \hline
            & \textbf{P}    & \textbf{R}    & \textbf{F}    \\ \hline
NC-Spatial  &  90.7         & 95.8          & 93.2          \\ 
NP-Spatial  &  97.1         & 92.9          & 94.9          \\ 
ENE-Spatial &  91.4         & 93.0          & 92.2          \\
Relation    &  91.0         & 93.6          & 92.3          \\ 
Latlong     &  98.4         & 98.6          & 98.5          \\

NC-Person   & 73.2          & 81.3          & 77.1          \\ 
NP-Person   & 88.3          & 89.2          & 88.7          \\
ENE-Person  & 87.5          & 73.9          & 80.1          \\

NP-Misc     & 74.5          & 80.0          & 77.1          \\ 
ENE-Misc    & 48.7          & 45.7          & 47.1          \\

Head        & 96.4          & 94.1          & 95.2          \\ 
Domain-mark & 98.2          & 98.5          & 98.3          \\ 
\hline
Micro Avg   &  91.5         & 92.1          & 91.8          \\ 
Macro Avg   &  86.3         & 86.4          & 86.2          \\\hline
    \end{tabular}
    \caption{Precision, recall and F1-scores evaluation for our fine-tuned Flair model.}
    \label{table:flair_scores}
\end{table}

\subsection{Few-shot prompting with generative LLMs}

\begin{figure*}[ht]
    \centering
    \includegraphics[width=\textwidth]{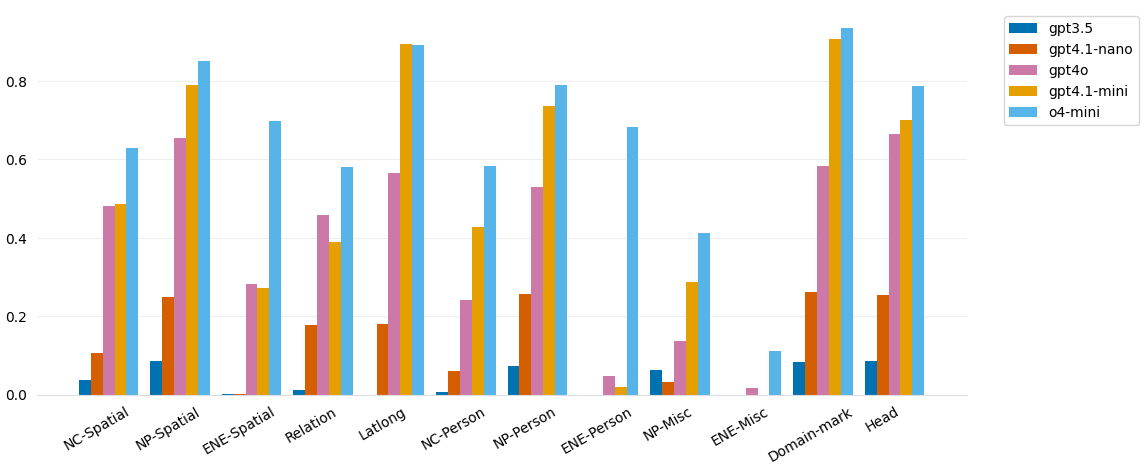}
    \caption{F1-scores comparison for OpenAI's GPT models.}
    \label{fig:gpt_models}
\end{figure*}

We evaluated the potential of few-shot prompting with state-of-the-art generative language models for the entity recognition task. In contrast to encoder-based fine-tuning approaches, this setup does not involve any training, relying solely on in-context learning. We focused our experiments on OpenAI’s models: GPT3.5, GPT-4o, GPT-4.1-mini, GPT-4.1-nano and o4-mini, using prompt templates with only 1 labeled example and a consistent JSON structured output format \cite{zeghidi2024evaluating}.

Figure~\ref{fig:gpt_models} shows the f1-scores comparison for each GPT models and each label. Table~\ref{table:gpt_scores} also reports precision and recall scores for the three best models: GPT-4o, GPT-4.1-mini and o4-mini.
These models demonstrate surprisingly strong performance, particularly GPT-o4-mini, which achieves the highest recall and competitive precision across most categories. In contrast, GPT-4o attains the best precision on several tags, including `NC-Spatial`, `NP-Spatial`, and `Relation`, but at the cost of significantly lower recall. GPT-4.1-mini shows more balanced scores but trails slightly behind GPT-o4-mini in both micro and macro averages.

Performance varies considerably across entity types. For well-represented and structurally clearer categories such as `Latlong`, `NP-Spatial`, `NP-Person`, and `Domain-mark`, models perform very well, even surpassing our fine-tuned spaCy NER for some labels such as `LatLong` and `Relation`. However, for more ambiguous or sparse classes like `ENE-Person`, `ENE-Misc`, and `NP-Misc`, results degrade substantially, indicating challenges in generalizing from few-shot context to rarer or noisier labels. Still, it is interesting to note that recall for `ENE-Person` improves dramatically from 2.5 with GPT-4o and 1.0 with GPT-4.1-mini to 57.8 with GPT-o4-mini, highlighting the rapid progress in newer model variants.

Notably, open-weights LLMs such as Llama 3-70b and local LLMs such as Mistral 7B were also evaluated but yielded consistently very poor performance and are therefore excluded from this study. Their limitations likely stem from insufficient instruction tuning and difficulty in following structured token-level labeling prompts.
Code and detailed results are available on Github\footnote{\url{https://github.com/GEODE-project/ner-llm}}.

\begin{table}[t]
    \centering
    \small
    \begin{tabular}{lcccccc}
        \hline
            & \multicolumn{2}{c}{\textbf{4o}} & \multicolumn{2}{c}{\textbf{4.1-mini}} & \multicolumn{2}{c}{\textbf{o4-mini}}  \\ \hline
            & \textbf{P} & \textbf{R} & \textbf{P} & \textbf{R} & \textbf{P} & \textbf{R}  \\ \hline
NC-Spatial  & \textbf{79.7}  & 34.5      & 40.9  & \textbf{59.8}   & 68.0  &  58.3 \\ 
NP-Spatial  & \textbf{93.3}  & 50.3      & 92.0   & 69.1  & 92.1  &  \textbf{79.1} \\ 
ENE-Spatial & 59.1  & 36.1      & \textbf{86.7}  &  16.2  & 81.3 & \textbf{61.2}  \\
Relation    & \textbf{62.5}  & \textbf{92.7}      & 26.9  &  69.7  & 46.8  & 76.3  \\ 
Latlong     & \textbf{92.5}  & 40.8      & 92.3  &  \textbf{86.6}  & \textbf{92.5}  & 85.8  \\

NC-Person   & \textbf{82.1}  & 14.2      & 36.4  & 52.0   & 56.2  & \textbf{60.4}  \\ 
NP-Person   & 74.8  & 40.9      & 78.3  & 69.5   & \textbf{82.7}  & \textbf{75.4}   \\
ENE-Person  & 45.5  & 2.5      & 66.7   & 1.0   & \textbf{83.3}  & \textbf{57.8} \\

NP-Misc     & 18.8  & 10.9      & 24.6  & 34.9  & \textbf{39.7}  & \textbf{42.9}  \\ 
ENE-Misc    & 2.9  & 1.2      & 0.00  & 0.00   & \textbf{15.9}  & \textbf{8.6}  \\

Head        & \textbf{92.3}  & 52.0      & 82.8  & 60.6   & 91.2  & \textbf{69.3}  \\ 
Domain-mark & \textbf{100}  & 41.1      & 92.6   & 88.8  & 96.5  & \textbf{90.6}  \\ 
\hline
Micro Avg   & \textbf{77.5}  & 33.4      & 58.5  & 57.4  & 75.6  & \textbf{70.5}   \\ 
Macro Avg   & 66.9  & 28.6      & 60.0  & 50.7  & \textbf{70.5}  & \textbf{63.8}   \\\hline
    \end{tabular}
    \caption{Precision and recall scores evaluation for OpenAI's GPT models.}
    \label{table:gpt_scores}
\end{table}

\section{Discussion}

\begin{table*}[ht]
\centering
 \small
\begin{tabular}{lccccccc}
\hline
\multirow{2}{*}{\textbf{Tag}}  & \multirow{2}{*}{\textbf{CRF}} & \textbf{Fine-tuned} & \textbf{Fine-tuned}   & \textbf{Fine-tuned}& \textbf{Fine-tuned} & \multirow{2}{*}{\textbf{GPT 4o-mini}}   \\ 
              &           & \textbf{spaCy NER}& \textbf{spaCy Spancat}  & \textbf{CamemBERT}& \textbf{Flair}     &  \textbf{}  \\  \hline
NC-Spatial    & 89.8           & 86.2           & \textbf{95.0}    & 93.4            & 93.2            & 62.8\\ 
NP-Spatial    & 90.9           & 83.0           & 93.5             & 94.7            & \textbf{94.9}   & 85.1\\ 
ENE-Spatial   & 89.9           & 88.3           & \textbf{93.4}    & 92.6            & 92.2            & 69.8\\
Relation      & 92.7           & 84.8           & 50.6             & 92.2            & \textbf{92.3}   & 58.0\\ 
Latlong       & 96.9           & 95.9           & 0.00             & 97.0            & \textbf{98.5}   & 89.0\\ 

NC-Person     & 64.0           & 61.9           & \textbf{77.5}    & 71.6            & 77.1            & 58.2\\ 
NP-Person     & 79.0           & 67.3           & \textbf{92.0}    & 87.4            & 88.7            & 58.2\\ 
ENE-Person    & 73.3           & 65.5           & 87.1             & \textbf{87.5}   & 80.1            & 68.2\\ 

NP-Misc       & 67.1           & 61.2           & 69.8             & 72.7            & \textbf{77.1}   & 41.2\\ 
ENE-Misc      & 43.9           & 32.5           & 0.00             & \textbf{52.2}   & 47.1            & 11.2\\ 

Head          & 88.1           & 85.6           & 45.1             & \textbf{96.8}   & 95.2            & 78.7\\ 
Domain-mark   & 99.0           & 96.2           & 95.8             & \textbf{99.1}   & 98.3            &  93.4\\  \hline

Micro avg     & 88.6           & 83.2           & 76.8             &  91.2           & \textbf{91.8}   & 73.0\\
Macro avg     & 81.2           & 75.7           & 66.6             &  \textbf{86.4}  & 86.2            & 66.2\\ \hline
\end{tabular}
\caption{Token-based F1-scores}
\label{table:summary_results}
\end{table*}

The results of our experiments are summarized in Table~\ref{table:summary_results} (only F1-scores are reported). 
Our evaluation of multiple NER architectures on the GeoEDdA corpus offers several insights into the strengths, limitations, and trade-offs inherent in applying modern NLP techniques to historical French texts. First, the performance gap between token-level and span-level models underscores the importance of selecting an appropriate task formulation for the complexity of the data. Flat models, while simpler to train and evaluate, often fail to capture overlapping and nested entity structures that are commonplace in early modern writing. Span-based models such as spaCy Spancat and multi-label transformer variants better handle such complexities but at the cost of higher computational overhead and sensitivity to class imbalance.

Transformer-based models like CamemBERT consistently outperformed traditional approaches across most entity types, particularly in the presence of overlapping spans. The joint-label strategy, despite simplifying the task, performed well for high-frequency entities but introduced combinatorial label sparsity for less common categories. Conversely, the multi-label formulation showed stronger recall for underrepresented and ambiguous classes, highlighting its ability to model overlapping annotations more flexibly.

The results from few-shot prompting with generative LLMs (e.g., GPT-4o and GPT-o4-mini) are particularly noteworthy. Despite receiving minimal contextual supervision, these models produced competitive results for frequent entity types and even demonstrated surprising generalization capabilities on some nested entities. However, performance remains highly variable across entity categories, with notable weaknesses for rare and context-dependent labels such as ‘ENE-Misc’ and ‘NP-Misc’. These findings point to both the potential and the current limitations of generative models for structured prediction tasks, especially when fine control over outputs and class consistency is required.

Our experiments also show that model performance is tightly linked to entity frequency and annotation structure. Entities such as 'Latlong' and 'Domain-mark', which have clearer boundaries and more consistent patterns, are reliably extracted even by lightweight models. In contrast, entities with broader semantic scope or ambiguous context (such as ‘NC-Person’ or ‘ENE-Misc’) remain challenging across all architectures. This suggests a need for more targeted modeling strategies, possibly involving entity-type-specific encoders or adaptive training techniques.

Finally, model interpretability and integration into downstream digital humanities workflows remain critical concerns. While state-of-the-art models achieve strong F1 scores, their opaque decision-making and high resource requirements can hinder adoption by historians and domain experts. Here, hybrid systems that combine neural predictions with symbolic post-processing (e.g., grammar-based filters, expert rules) may offer a viable path forward, allowing for more controlled and interpretable results.

\section{Conclusion}

Named Entity Recognition in historical texts remains a complex and multifaceted problem due to the inherent linguistic variability and nested entity structures prevalent in early modern corpora. Our systematic benchmarking across six diverse NER architectures using the GeoEDdA dataset reveals that transformer-based models, particularly CamemBERT fine-tuned with joint-label or multi-label classification strategies, achieve the highest overall performance, especially on challenging nested entities. Lightweight models such as Flair and spaCy Spancat also show competitive results, particularly for person entities, demonstrating their utility in resource-constrained environments.

Importantly, few-shot prompting with generative GPT models emerges as a promising alternative for low-resource or zero-shot settings, delivering competitive recall on well-represented entity types and offering a rapid pathway to bootstrap annotations in active learning workflows. Nevertheless, their performance remains inconsistent for rare or highly ambiguous classes compared to fine-tuned discriminative models.

Our findings underscore the ongoing need for hybrid approaches that integrate symbolic rule-based reasoning with neural models to leverage domain expertise and address the linguistic intricacies of historical French texts. Future work should explore specialized pretraining on historical corpora, richer annotation schemes, and hybrid modeling frameworks that combine the strengths of deep learning with expert linguistic knowledge. These advances will be critical to enhancing downstream tasks such as geographic disambiguation, historical place-name mapping, and reconstruction of intellectual networks from digitized archival materials.

\section*{Acknowledgments}

The authors are grateful to the ASLAN project (ANR-10-LABX-0081) of the Université de Lyon, for its financial support within the French program "Investments for the Future" operated by the National Research Agency (ANR).

\bibliography{acl_latex}

\end{document}